\documentclass[11pt]{article}
\usepackage{geometry}
\usepackage{coling2020}
\usepackage{times}
\usepackage{url}
\usepackage{latexsym}
\usepackage{microtype}
\usepackage{graphicx}
\usepackage{url}
\usepackage{float}
\usepackage{makecell}
\usepackage[utf8]{inputenc}
\usepackage[arabic, english]{babel}
\usepackage{adjustbox}
\usepackage{comment}

\colingfinalcopy 

\title{Is this sentence valid? An Arabic Dataset for Commonsense Validation}

\author{Saja Tawalbeh, Mohammad AL-Smadi\\Computer Science Department\\ 
Jordan University of Science and Technology\\ 
P.O.Box: 3030 Irbid 22110, Jordan. \\
{\tt sktawalbeh16@cit.just.edu.jo, masmadi@just.edu.jo}}

\providecommand{\keywords}[1]{\textbf{\textit{Index terms---}} #1}

\date{}

\begin{document}
\maketitle
\begin{abstract}
The commonsense understanding and validation remains a challenging task in the field of natural language understanding. Therefore, several research papers have been published that studied the capability of proposed systems to evaluate the models ability to validate commonsense in text. In this paper, we present a benchmark Arabic dataset for commonsense understanding and validation as well as a baseline research and models trained using the same dataset. To the best of our knowledge, this dataset is considered as the first in the field of Arabic text commonsense validation. The dataset is distributed under the Creative Commons BY-SA 4.0 license and can be found on GitHub\footnote{Arabic dataset for commonsense validation, \url{https://github.com/msmadi/Arabic-Dataset-for-Commonsense-Validationion}}.

\end{abstract}

\section{Introduction}
\label{intro}
Text understanding and validation demands the ability to understand between lines instead of the general idea. The essential difference between the human being and a machine regarding text understanding is that humans have the ability to process and validate textual data and make sense out of it. Consequently,  the essential question is how to increase the ability of the machine to have the ability of commonsense understanding. 
Therefore, introducing commonsense to natural language understanding is considered as a qualitative change of research to investigate the machine ability in solving commonsense problems. Commonsense, in general, represents the aspect of the relation semantic properties and a piece of text \cite{asher1995toward}. \\

In the past fifty years of AI research, automated commonsense reasoning has barely progressed in the field of NLP research \cite{davis2004introduction}. In NLP, using transfer learning models \cite{yosinski2014transferable} as well as introducing transformers \cite{vaswani2017attention} has revolutionized the research field and increased the ability of machines in text commonsense validation and understanding. Furthermore, modern language models such as BERT, ELMo has enhanced machines' performance make-sense, validate and understand text \cite{davis2017logical}.\\ 

The existing dataset, covers commonsense knowledge indirectly. However, recently, \cite{wang2019does,wang2020semeval} presented a dataset in English that consists of two sentences, the machine should determine which one is against commonsense, then should explain why the sentence against commonsense, and finally, should generate the reason automatically.\\

To the best of our knowledge, there is no publicly available dataset in Arabic for commonsense validation and understanding of text. Therefore, in this paper present an Arabic dataset for commonsense validation. Moreover, we trained popular transformers i.e. BERT, USE, and ULMFit to provide a baseline model for the presented dataset.  The models' task is then to validate two natural language sentences of similar wording and to identify the one that does not make-sense. \\

\section{Related Work}
\label{related}
The natural language understanding frameworks are getting more attention after commonsense studies were conducted \cite{wang2018modeling}. Several datasets were prepared such as the Choice of Plausible Alternatives (COPA) \cite{roemmele2011choice} which focuses on events and consequences, in which each question given two alternatives to recognize the appropriate cause or result of the premise. Other datasets inspired by COPA was released, for instance,  JHU Ordinal Commonsense Inference (JOCI) dataset \cite{zhang2017ordinal} which has five labels starting from 5 (very likely) to 1 (impossible), tends to identify the plausibility of human response after a certain situation. Moreover, Situations with Adversarial Generations (SWAG) \cite{zellers2018swag} is a large-scale adversarial dataset for a grounded commonsense inference that is used to determine the action that probably happens after a specific situation. The main intention of SWAG dataset is to focus on the pre-situations and/or the after-situations of certain situations without looking to the reasons why they occur. The Story Cloze Test and ROCStories Corpora provided by \cite{mostafazadeh2016corpus,sharma2018tackling} inspired by grasp reading and text material questions are used to identify the appropriate answer from the provided materials. Ostermann, 2018 proposed a dataset for narrative text, which tells about several type of questions with two candidate answers \cite{ostermann2018mcscript}. \\

Other Question Answering (QA) datasets for factual commonsense
knowledge reasoning were released. For instance, SQUABU \cite{davis2016write} presented a simple test of commonsense and scientific questions. Moreover, CommonsenseQA \cite{talmor2018commonsenseqa} provided a ConceptNet that has been used by crowd workers to create questions for the presented dataset. Similarly, BookQA \cite{mihaylov2018can} presented questions and their candidate answers. COSMOS QA \cite{huang2019cosmos} is large-scale dataset contains multiple-choice questions with four answers. \\

Recently, Wang, 2020  proposed a multi-task \cite{wang2019does,wang2020semeval} aiming to recognize the hidden inference of the commonsense facts using SemEval commonsense validation and explanation dataset as well as performing RoBERTa to solve the explanation task \cite{wang2020cuhk}. Similarly, \cite{saeedi2020cs} proposed a model using RoBERTa to solve the SemEval commonsense validation and explanation task, they conducted this research by reframing the classification task to multiple-choice task aims to increase the system's performance.

\section{Dataset Collection and Translation}

The Arabic dataset for commonsense validation is based on the  Commonsense Validation and Explanation (ComVE) task which is inspired by the work of \cite{wang2019does}. The original dataset \footnote{ \url{https://github.com/wangcunxiang/SemEval2020-Task4-Commonsense-Validation-and-Explanation}} was prepared using Amazon Mechanical Turk where crowds were asked to write a sensible sentence and non-sensible sentences as (s1 and s2). For the sake of this research, we translated the original English dataset for commonsense validation. Each example in the provided dataset is composed of 2 sentences: \{s1, s2\} and a label indicating which one is invalid. The pair of sentences are too similar with a slight difference between there words to identify which sentence does not make sense.\\

To the best of our knowledge, there is no Arabic dataset publicly available for commonsense validation. The provided dataset has 12k rows and consists of three files: train, validation, and test file. Each row consists of two sentences and the label of the non-sensible sentence. Table \ref{distrbution} shows the distribution ans size of the dataset. Whereas, Table \ref{dataset_EXAMPLE} presents examples from the Arabic dataset for the sense-making task.

\begin{table}[h!]
\begin{center}
\begin{tabular}{|l|l|l|l|l|}
\hline \bf File & \bf Train & \bf Validation & \bf Test & \bf Total \\ \hline
\# of rows & 10000 & 1000 & 1000 & 12000 \\

\hline
\end{tabular}
\end{center}
\caption{\label{distrbution}The Dataset distribution}
\end{table}

\begin{table*}[h!]
\centering
\begin{tabular}{lccc}
  id & Sent0 & Sent1 & label\\
  \hline
   26 & \makecell{\textAR{أنا ألبس فيل} \\ I dress an elephant} & \makecell{\textAR{أنا ألبس بنفسي} \\ I dress myself} & 0 \\ \hline
   
   34 & \makecell{\textAR{قرأت النجوم} \\ I read stars} & \makecell{\textAR{قرأتني النجوم} \\ Stars read me} & 1 \\ \hline
   
   4271 & \makecell{\textAR{كرة القدم مربعة} \\ Football is square} & \makecell{\textAR{كرة القدم مستديرة} \\ Football is round} & 0 \\ \hline
   
   5805 & \makecell{\textAR{الأرض أصغر من الشمس} \\ the earth is smaller than the sun} & \makecell{\textAR{الأرض أصغر من القمر} \\ the earth is smaller than the moon} & 1 \\ \hline
   
   9231 & \makecell{\textAR{تطير الطيور في السماء} \\ birds fly in the sky} & \makecell{\textAR{يطير البشر في السماء} \\ humans fly in the sky} & 1 \\ \hline
   
\end{tabular}
\caption{\label{dataset_EXAMPLE} Examples from the provided Arabic dataset with the English original ones. The label indicates the id of the invalid sentence (i.e. 0: for the first sentence or 1: for the second sentence)}
\end{table*}

\section{Experimentation \& Results}
As baseline research, several state-of-the-art models i.e. BERT \cite{devlin2018bert}, USE \cite{cer2018universal}, and ULMFit \cite{howard2018universal} have been trained using the proposed Arabic dataset to make-sense of the provided couple of sentences. Trained models were evaluated based on their performance accuracy.\\

As shown in Table \ref{results}, BERT has a significant advantage compared to ULMFiT and USE. Results go in line with previous research where BERT has achieved 70.1\% accuracy for the original English dataset \cite{wang2019does} overcoming other models such as ULMFit. 

\begin{table}[h!]
\begin{center}
\begin{tabular}{|c|c|}
\hline \bf Model & \bf Accuracy \\ \hline
USE & 48.3\\
ULMFiT  & 50.5 \\ 
BERT  & 62.7\%\\ \hline
\end{tabular}
\end{center}
\caption{\label{results} The experimentation results for the trained baseline models}
\end{table}

\section{Conclusion}
This paper presents a benchmark dataset for Arabic text commonsense and validation. Presented dataset were used to train baseline models such as BERT \cite{devlin2018bert}, USE \cite{cer2018universal}, and ULMFit \cite{howard2018universal}. Evaluation results show that BERT outperformed the other two models and achieved a 62.7\% in making-sense and validation input sentences.

\section*{Acknowledgments}
This research is partially funded by Jordan University of Science and Technology, Research Grant Number: 20170107.

\bibliographystyle{coling}
\bibliography{semeval2020}

\end{document}